\documentclass{article}
\usepackage{graphicx}
\usepackage[below]{placeins}
\renewcommand{\(}{\left(}
\renewcommand{\)}{\right)}
\newcommand{\beq}{\begin{equation}}
\newcommand{\eeq}{\end{equation}}
\newcommand{\beqa}{\begin{eqnarray}}
\newcommand{\eeqa}{\end{eqnarray}}
\newcommand{\avsi}{\langle s_i \rangle}
\newcommand{\sets}{\mathcal{S}}
\newcommand{\setv}{\mathcal{V}}
\newcommand{\setsi}{\mathcal{S}^{(i)}}
\newcommand{\setvi}{\mathcal{V}^{(i)}}

\begin{document}
\begin{titlepage}
\begin{flushright}
LU TP 98-29\\
\today \\
\end{flushright}
\vspace{0.4in}
\LARGE
\begin{center}

{\bf An Efficient Mean Field Approach\\ to the Set Covering Problem}\\
\vspace{0.3in}
\large
Mattias Ohlsson\footnote{mattias@thep.lu.se},
Carsten Peterson\footnote{carsten@thep.lu.se} and
Bo S\"oderberg\footnote{bs@thep.lu.se}\\
\vspace{0.1in}
Complex Systems Group, Department of Theoretical Physics\\
University of Lund, S\"olvegatan 14A, S-223 62 Lund, Sweden\\
{\tt http://www.thep.lu.se/tf2/complex/}
\vspace{0.3in}

Submitted to {\it European Journal of Operations Research}

\end{center}
\vspace{0.2in}
\normalsize

\textbf{Abstract}

A mean field feedback artificial neural network algorithm is developed
and explored for the set covering problem. A convenient encoding of
the inequality constraints is achieved by means of a multilinear
penalty function. An approximate energy minimum is obtained by
iterating a set of mean field equations, in combination with
annealing. The approach is numerically tested against a set of
publicly available test problems with sizes ranging up to $5 \times
10^3$ rows and $10^6$ columns. When comparing the performance with
exact results for sizes where these are available, the approach yields
results within a few percent from the optimal solutions. Comparisons
with other approximate methods also come out well, in particular given
the very low CPU consumption required -- typically a few seconds.
Arbitrary problems can be processed using the algorithm via a public
domain server.

\textit{Keywords:}
Neural networks; Mean field annealing; Set covering; 
Combinatorial optimization
\end{titlepage}

\normalsize
\section{Introduction}

The {\it set covering problem} ({\bf SCP}) is a well known NP-hard
combinatorial optimization problem, which represents many real-world
resource allocation problems. Exact solutions can be obtained by
e.g. a branch-and-bound approach for modestly sized problems. For
larger problems various approximative schemes have been suggested (see
e.g.  \cite{beas1, beas2}). In this paper we develop a novel approach
based on feedback Artificial Neural Networks ({\bf ANN}), derived from
the mean field approximation to the thermodynamics of spin systems.

ANN is a computer paradigm that has gained a lot of attention during
the last 5-10 years. Most of the activities have been directed towards
feed-forward architectures for pattern recognition or function
approximation. ANN, in particular feedback networks, can also be used
for difficult combinatorial optimization problems
(e.g. \cite{tank}-\cite{lager2}).  Here ANN
introduces a new method that, in contrast to most existing search and
heuristics techniques, is not based on exploratory search to find the
optimal configuration. Rather, the neural units find their way in a
fuzzy manner through an interpolating, continuous space towards good
solutions. There is a close connection between feedback ANN and spin
systems in statistical physics.  Consequently, many mathematical tools
used for dealing with spin systems can be applied to feedback ANN. Two
steps are involved when using ANN for combinatorial optimization:
\begin{enumerate}
\item Map the problem onto an energy function, e.g.
\beq
\label{E}
        E\(\sets\) =\frac{1}{2} \sum_{ij} w_{ij} s_i s_j
\eeq
where $\sets=\left\{s_i;i=1\dots N\right\}$ is a set of binary
spin variables $s_i\in\{0,1\}$, representing the elementary choices
involved in minimizing $E$, while the {\it weights} $w_{ij}$ encode
the costs and constraints.
\item To find configurations with low $E$, iterate the
mean field ({\bf MF}) equations
\beq
\label{mf}
        v_i =
        1 \left/ \(1+\exp\(\frac{1}{T}\frac{\partial E\(\setv\)}{\partial v_i}\)\)\right.
\eeq
where $T$ is a fictitious temperature while $\setv =
\left\{v_i\right\}$, where $v_i\in [0,1]$ represents the thermal
average $\langle s_i\rangle_T$, and allows for a probabilistic
interpretation.
\end{enumerate}

Eqs. (\ref{E},\ref{mf}) only represent one example. More elaborate
encodings have been considered, e.g. based on {\it Potts} spins
allowing for more general basic decisions elements than simple binary
ones \cite{pet1}. A propagator formalism based on Potts neurons has
been developed for handling topological complications in e.g. routing
problems \cite{lager}\cite{lager1}.

The ANN approach for SCP that we develop here differs from the one
that was successfully used for the somewhat related knapsack problem
in \cite{ohl,mo1}, in particular with respect to encoding the
constraints. Whereas a non-linear step-function was used in
\cite{ohl,mo1}, we will here use a multilinear penalty, which in
addition to being theoretically more appealing, also appears to be
very efficient. Furthermore, an automatic procedure for setting the
relevant $T$-scale is devised.

The algorithm is extensively tested against a set of publicly
available benchmark problems \cite{OR} with sizes
(rows$\times$columns) ranging from 200$\times$1000 to
5000$\times$10$^6$. The approach yields results, typically within a
few percent from the exact optimal solutions, for sizes where these
are available. Comparisons with other approximate methods also come
out well. The algorithm is extremely rapid -- the typical CPU demand
is only a few seconds (on a 400MHz Pentium II).

A public domain WWW server has been set up, where arbitrary
problems can be solved interactively.

This paper is organized as follows: In Sect. 2 we define the set
covering problem, and in Sect. 3 we describe its encoding in terms of
a neural network energy function and discuss the mean field
treatment. Sect. 4 contains numerical explorations and comparisons. A
brief summary is given in Sect. 5. Appendices A and B contain
a derivation of the mean field equations, and some algorithmic implementation
issues, respectively. Tables from the numerical explorations are found
in Appendix C, while Appendix D contains pointers and instructions for
the WWW server. It should be stressed that this paper is
self-contained -- no prior knowledge of feedback neural networks or
the mean field approximation is necessary.

\section{The Set Covering Problem}

The set covering problem (SCP) is the problem of finding a subset of
the columns of an $M$x$N$ \textit{zero-one} matrix $A = \{a_{ki} \in
\{0,1\}; k=1,\dots,M; i=1,\dots,N\}$ that covers all rows at a minimum
cost, based on a set of column costs $\{c_i; i=1,\dots,N\}$. SCP is
conveniently described using a set of binary variables, $\sets = \{s_i
\in \{0,1\}; i=1,\dots,N\}$. More precisely, SCP is defined as
follows:
\beqa
\label{def}
        \textrm{Minimize} \qquad && \sum_{i=1}^N c_i s_i  \\
\label{constr1}
        \textrm{subject to} \qquad && \sum_{i=1}^N a_{ki} s_i  \geq 1, \qquad k=1,\dots,M \\
\label{constr2}
        \textrm{with} \qquad && s_i \in \{0,1\} \qquad i=1,\dots,N
\eeqa
Eq. (\ref{constr1}) states that at least one column must cover a
particular row. The special case where all costs $c_i$ are equal is
called the {\it unicost} SCP.  There is subclass of SCPs that has a
nice graphical interpretation: If the \textit{zero-one} matrix $A$ has
the property that each row contains exactly two 1's then we can
interpret $A$ as the vertex-edge matrix of a graph with $N$ vertices
and $M$ edges. The unicost SCP is then a vertex covering problem,
where the task is to find the minimal number of vertices that covers
all edges of the graph. SCPs (including weighted vertex covering) are
NP-hard combinatorial optimization problems. If the inequalities of
Eq. (\ref{constr1}) are replaced by equalities, one has the \textit{set
partition problem} ({\bf SPP}). Both SCP and SPP have numerous
resource allocation applications.

The following quantities will be used later:
\beqa
\label{rho}
        \textrm{Density:} \qquad && \rho = \frac{1}{MN} \sum_{ki} a_{ki} \\
\label{colsum}
        \textrm{Column Sums:} \qquad && \hat{N}_i = \sum_k^M a_{ki} \\
\label{rowsum}
        \textrm{Row Sums:} \qquad && \hat{M}_k = \sum_i^N a_{ki} 
\eeqa
%

\section{The Mean Field Approach}

\subsection{The Energy Function for SCP}

We start by mapping the SC problem of Eqs. (\ref{def}, \ref{constr1},
\ref{constr2}) onto a spin energy function $E(\sets)$ (step 1 in the
introduction),
\beq
\label{En_s}
        E(\sets) = \sum_{i=1}^N c_i s_i + \alpha \sum_{k=1}^M
        \prod_{i=1}^N \( 1 - a_{ki} s_i \)
\eeq
The first term yields the total cost and the second one represents the
covering constraint of Eq. (\ref{constr1}) by imposing a penalty if a
row is not covered by any column.

The constraint term is a multilinear polynomial in the spin variables
$s_i$, i.e.  it is a linear combination of products
$s_{1}s_{2}\dots s_{K}$ of distinct spins.  This is attractive from a
theoretical point of view, and the differentiability of $E$ enables a
more quantitative analysis of the dynamics of the
mean field algorithm.

An alternative would be to implement the inequality constraints using a
piecewise linear function \cite{ohl},
\beq
        \alpha \sum_{k=1}^M \phi\( 1 - \sum_{i=1}^N a_{ki} s_i \)
\eeq
with $\phi(x) = x\Theta(x) = x$ if $x>0$ and 0 otherwise. This yields
a non-differentiable energy function and generally an inferior
performance as compared to the polynomial representation (\ref{En_s}).

\subsection{The Statistical Mechanics Framework}

The next step is to minimize $E(\sets)$. Using some local updating
rule will most often yield a local minimum close to the starting
point, with poor solutions as a result. Simulated annealing ({\bf SA})
\cite{kirk} is one way of escaping from local minima since it allows
for uphill moves in $E$. In SA a sequence of configurations $\sets$ is
generated according to a stochastic algorithm, such as to emulate the
probability distribution
\beq
        P(\sets) = \frac{e^{-E\(\sets\)/T}}
        {\displaystyle{\sum_{\sets'}} e^{-E\(\sets'\)/T}}
\eeq
where the sum runs over all possible configurations $\sets'$. The
parameter $T$ (temperature) acts as a noise parameter. For large $T$
the system will fluctuate heavily since $P(\sets)$ is very flat. For
the SCP this implies that the sequence contains mostly poor and
infeasible solutions. On the other hand, for a small $T$, $P(\sets)$
will be narrow, and the sequence will be strongly dependent upon the
initial configuration and contain configurations only from a small
neighborhood around the initial point. In SA one generates
configurations while lowering $T$ (annealing), thereby diminishing the
risk of ending up in a suboptimal local minimum. This is quite
CPU-consuming, since one has to generate many configurations for each
temperature following a careful annealing schedule (typically $T_k =
T_0 / \log(1+k)$ for some $T_0$) in order to be certain to find the
global minimum.

In the {\it mean field} ({\bf MF}) approach the costly stochastic SA
is approximated by a deterministic process. MF also contains an
annealing procedure. The original binary variables $s_i$ are replaced
by continuous \textit{mean field} variables $v_i \in [0,1]$, with a
dynamics given by iteratively solving of the MF equations for each $T$.

An additional advantage of the MF approach is that the continuous mean
field variables can evolve in a space not accessible to the original
variables. The intermediate configurations at non-zero $T$ have a
natural probabilistic interpretation.

\subsection{Mean Field Theory Equations}

Our objective is now to minimize $E$ using the MF method. The binary
spin variables $s_i$ are replaced by mean field variables $v_i$,
representing solutions to the MF equations (a derivation of these is
given in Appendix A).
\beq
        v_i =
        1\left/\(1 + \exp\(\frac{1}{T}\Delta E_i\(\setvi\)\)\)\right.
        ,\quad i=1,\dots,N
\label{update}
\eeq
where
\beq
        \Delta E_i\(\setvi\)
        = E\(\setvi,v_i=1\) - E\(\setvi,v_i=0\)
        \quad = \quad \partial E/\partial v_i
\label{deltaE}
\eeq
The set $\setvi$ denotes the complementary set $\{v_j, j\neq i\}$.
From Eq.  (\ref{En_s}) we get
\beq
\label{dE}
        \Delta E_i\(\setvi\) = c_i - \alpha\sum_{k=1}^M a_{ki}
        \prod_{j\neq i}^N
        \( 1 - a_{kj} v_j \)
\eeq
The MF equations are solved iteratively while annealing in $T$. It
should be noted that the equation for $v_i$ does not contain any
feedback, i.e no explicit dependence on $v_i$ itself. This yields a
smooth convergence, and it is usually only necessary with a few
iterations at each value of $T$. What remains to be specified are the
parameters $\alpha$ and $T$.  The latter will be discussed next and we
return to the choice of $\alpha$ in connection with the numerical
explorations in Sect. 4.

\subsection{Critical Temperatures}

In the limit of high temperatures, $T\to \infty$, the MF variables
$\{v_i\}$ will, under the dynamics defined by iteration of
Eq. (\ref{update}), converge to a trivial symmetric fixed point with
$v_i = 1/(1+\exp(0)) = \frac{1}{2}$,
corresponding to no decision taken. At a finite but high $T$, the
corresponding fixed point will typically deviate slightly from the
symmetric point.

For many problems, a bifurcation occurs (indicative of a transition
from a disordered phase to an ordered one) at a critical temperature
$T_c$, where the trivial fixed point loses stability and other fixed
points emerge, which as $T\to 0$ converge towards definitive candidate
solutions to the problem, in terms of $v_i \in \{0,1\}$.  For some
problems, a cascade of bifurcations occur, each at a distinct critical
temperature, but in the typical case there is a single bifurcation.

It is then of interest to estimate of the position of $T_c$, which
defines a suitable starting point for the MF algorithm.
Such an estimate can be obtained by means of a linear stability
analysis for the dynamics close to the fixed point. For the special
case of a symmetric unicost SCP with constant row and column sums for
the matrix $A$, there will be a sharp transition around
\beq
        T_c \approx \alpha \rho^2 M 2^{-\rho N}
\eeq
where $\rho$ is defined in (\ref{rho}).

For non-unicost problems, $T_c$ is harder to estimate, and there might
even be no bifurcation at all. For such problems, a suitable initial
$T$ is instead determined by means of a fast preliminary run of the
algorithm (see below).

\section{Numerical Explorations}

\subsection{Implementation Details}

The annealing schedule for $T$ and the value of the constraint
parameter $\alpha$ have to be determined before we can run the
algorithm. The former is accomplished by a geometric decrease of $T$,
\beq
\label{T_update}
        T_{t+1} = k T_t
\eeq
where $k$ is set to 0.80 and $T_0$ is determined by a fast prerun of
the algorithm (see below). The number of iterations of Eqs.
(\ref{update}) for each value of $T$ is not fixed; $T$ is lowered only
when all $v_i$ have converged.

In order to ensure a valid solution at low $T$, the size of the
constraint term in Eq. (\ref{dE}) must be larger than the largest cost
$c_{max}$ that is part of the solution. Using a too large $\alpha$
will however reduce the solution quality since $E$ is then dominated
by the covering constraint. Our choice of $\alpha$ therefore depends
on $c_{max}$. Ideally, $M/(\rho M) = 1/\rho$ columns would suffice to
cover each row of $A$. If we further optimistically assume that the
$1/\rho$ smallest costs can be chosen for the solution, $c_{max}$ can
easily be found. However, for most problems we need more than $1/\rho$
columns, which makes it difficult to estimate $c_{max}$ except for
unicost problems where all column costs are equal.

To determine (an approximate) $c_{max}$ for a non-unicost SCP, we
perform a fast prerun with a smaller annealing factor $k = 0.65$, and
with $\alpha = 1.01$. From this prerun one can also obtain an estimate
the critical temperature $T_c$ as the $T$ where the saturation
(defined below) deviates from 0. The second run is then initated at
$T_0 = 2 \, T_c$, thereby avoiding unnecessary updates of $v_i$ at high
$T$.

This procedure for setting $\alpha$ also requires rescaling of the
costs; for all problems we set $c_i \to c_i / \max_j(c_j)$. See Table
\ref{param} for a summary of the parameters used.
\begin{table}[ht]
\begin{center}
\begin{tabular}{lccc}
\hline
& Unicost SCP & \multicolumn{2}{c}{Non-unicost SCP}  \\
& & Prerun & Second run \\ 
\hline
$k$      & 0.80 & 0.65  & 0.8                \\  
$\alpha$ & 0.5  & 1.01  & $1.05 * c_{max}^*$ \\
$T_0$    & 50   & 50    & $2 * T_c$          \\
\hline
\end{tabular}
\end{center}
\caption{Summary of the parameters $k$, $\alpha$ and $T_0$ used in the algorithm.}
\label{param}
\end{table}

The evolution of the MF variables $\{v_i\}$ is conveniently monitored
by the {\it saturation} $\Sigma$,
\beq
\label{sat}
        \Sigma = \frac{4}{N} \sum_{i=1}^N \( v_i - 1/2 \) ^2
\eeq
A completely ``undecided'' configuration, $\Sigma = 0$, means that
every $v_i$ has the value $1/2$. During the annealing process, as the
$v_i$ approach either 1 or 0, $\Sigma$ converges to 1. The transition
between $\Sigma=0$ and $\Sigma=1$ is usually smooth for a generic SCP.
However, when a bifurcation is encountered (see above), $\Sigma$ can
change abruptly; this occurs e.g. for unicost SCP where there is no
natural ordering among the costs $\{c_i\}$.

\begin{figure}[htb]
\begin{center}
\includegraphics[width=7cm,angle=-90]{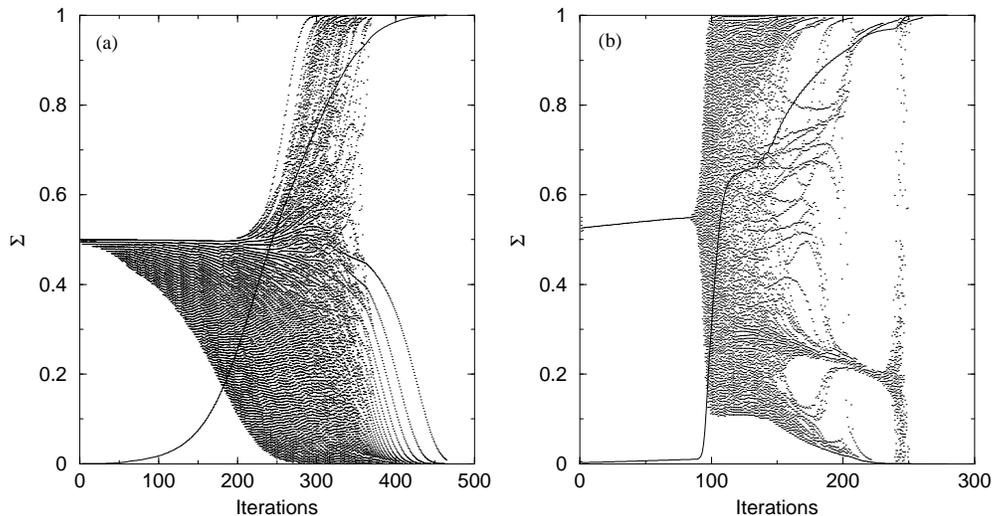}
\end{center}
\caption{Evolution of the mean field variables $v_i$ as $T$ is 
        lowered for {\tt 4.1} {\bf (a)} and {\tt cyc09} {\bf (b)}
        respectively. Also shown is $\Sigma$ (Eq. (\protect\ref{sat})). Note
        that the number of iterations is intentionally large for
        visualization purposes.}
\label{devel}
\end{figure}

Fig. \ref{devel} shows the evolution of the mean field variables
$\{v_i\}$ and the saturation $\Sigma$ for the problems {\tt 4.1} and
{\tt cyc09}; the latter is a unicost problem (see Appendix C), and as
can be seen from Fig. \ref{devel}b, it clearly exhibits a bifurcation.

A summary of the algorithm can be found in Appendix B, while Appendix D
gives the address of and instructions for a WWW server, where the MF
algorithm is applied to user-defined SCPs.

\subsection{Numerical Results}

The performance of the algorithm is evaluated using 16 problem sets
found in the OR-Library benchmark database \cite{OR}. These 16 sets
consist of 91 problems, out of which 19 are unicost SCP. The algorithm
is coded in C and the computations are done on a 400MHz Pentium II PC.
The details of the OR-library problems are given in Table \ref{pdef}
and our results can be found in Tables \ref{res1}, \ref{res2} and
\ref{res3} in Appendix C. The optimal or currently best known values
are taken from refs.  \cite{beas1,beas2,gross,res,sass1,sass2}.

For each of the test SCPs, 10 trials of our algorithm are performed.
In Tables \ref{res1}-\ref{res3}, the best and the average costs are
listed for each problem. For 10 of the problems our method found the
optimal solution. The MF results typically are within a few percent of
the optimal solutions, as can be seen in Table \ref{res4}. Large
relative deviations from optimum are seen in the unicost {\tt CLR}
problems, which appear to be difficult for our approach. However, the
optimal (integer) costs for these problems are low (23~-~26); this
gives a large effect on the relative deviations even for a small
change in the found costs. Decreasing the obtained costs by unity will
change the mean relative deviation from $16\%$ to $10\%$. It is also important
to notice that we use a common set of algorithm parameters ($\alpha,
k, T_0$) for all unicost problems, without any parameter optimization
for each problem. Another set of parameters might be more advantageous
for the {\tt CLR} problems.
\begin{table}[h]
\begin{center}
\begin{tabular}{l|cccccccc}
\hline
Problem set & 4 & 5 & 6 & A & B & C & D & E \\
Rel. deviation (\%) & 2.1 & 2.7 & 3.5 & 2.2 & 1.1 & 1.7 & 2.9 & 0 \\  
\hline\hline
Problem set & NRE & NRF & NRG & NRH & Rail & CYC & CLR & STS \\
Rel. deviation (\%) & 3.5 & 4.4 & 3.1 & 2.0 & 6.8 & 3.7 & 16 & 2.5 \\  
\hline
\end{tabular}
\end{center}
\caption{Mean relative deviation from the optimal (or best known) solution.}
\label{res4}
\end{table}

\begin{figure}[htb]
\begin{center}
\includegraphics[width=13cm]{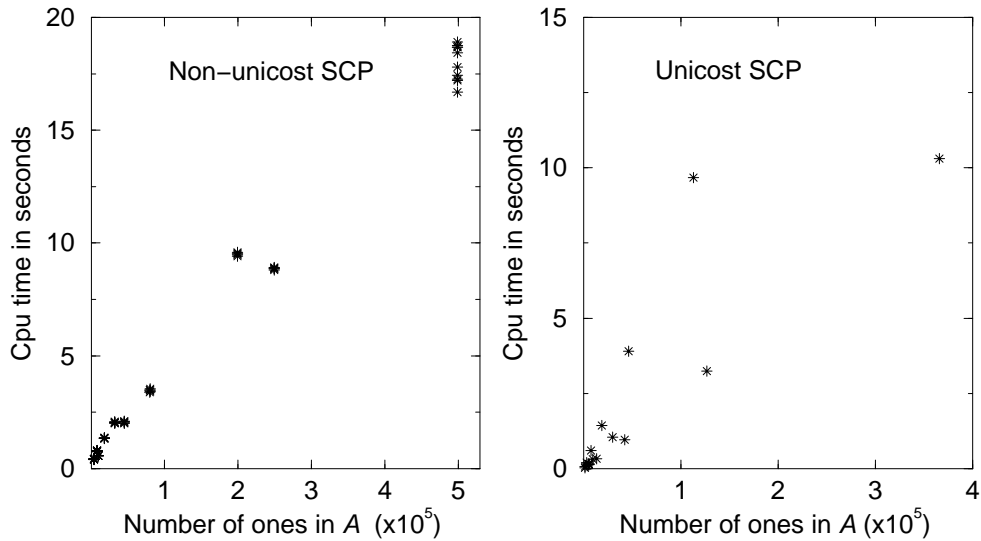}
\end{center}
\caption{Solution time in seconds, referring to a Pentium II 400 MHz computer, 
  as function of the number of non-zero entries in the problem matrix
  $A$. Left and right figure shows non-unicost and unicost SCP,
  respectively The rail problems are not shown since a slightly
  different implementation is used for these problems (see Appendix B
  for details).}
\label{cpu}
\end{figure}

In our implementation of the MF algorithm, the time $\tau$ for a
complete update of all variables $\{v_i\}$ scales approximately
linearly with the number of non-zero entries in $A$,
\beq
\label{tau}
        \tau \propto NM\rho
\eeq
This appealing property is feasible due to efficient calculations of
$\Delta E_i$ in Eq. (\ref{dE}) that utilizes the sparse nature of $A$.
For the total solution time, $\tau$ should be multiplied by the number
of iterations needed -- empirically around 100, independently of
problem size. Fig. \ref{cpu} shows the mean solution time versus
$NM\rho$.

Compared to other heuristic approaches, ours does not find the optimal
cost as often as e.g the genetic algorithm \cite{beas2}. It is however
very competitive with respect to speed. In ref. \cite{gross} nine
different approximation algorithms were tested on a large number of
unicost problems (including the set considered here); our approach is
comparable to the top ones in both performance and speed.

\section{Summary}

We have developed a mean field feedback neural network approach for
solving set covering problems. The method is applied to a standard set
of benchmark problems available in the OR-library database. The method
is also implemented in a public WWW server.

The inequality constraints involved are conveniently handled by means
of a multi-linear penalty function that fits nicely into the mean
field framework. The bifurcation structure of the mean field dynamics
involved in the neural network approach is analyzed by means of a
linearized dynamics. A simple and self-contained derivation of the
mean field equations is provided.
 
High quality solutions are consistently found throughout a range of
problem sizes ranging up to $5 \times 10^3$ rows and $10^6$ columns
for the OR-library problems without having to fine-tune the
parameters, with a time consumption scaling as the number of non-zero
matrix elements. The approach is extremely efficient, typically
requiring a few seconds on a Pentium II 400 MHz computer.

The mean field approach to SCP can easily be modified to apply to the
related, and more constrained, \textit{set partitioning problem}. One
simply has to replace the inequality constraint term with one that
handles the equality constraint present in the set partitioning
problem.

\subsection*{Acknowledgments:}

This work was in part supported by the Swedish Natural Science
Research Council, the Swedish Board for Industrial and Technical
Development and the Swedish Foundation for Strategic Research.


\newpage

\appendix
\renewcommand{\thesection}{Appendix \Alph{section}.
\setcounter{equation}{0}}
\setcounter{table}{0}
\renewcommand{\theequation}{\Alph{section}\arabic{equation}}
\renewcommand{\thetable}{\Alph{section}\arabic{table}}
\newcommand{\app}[1]{\newpage\section{#1}}

\app{Mean Field Approximation}

Here follows for completeness a derivation of the mean field equations
(see e.g. \cite{parisi}). Let $E\(\sets\)$ be an energy function of a
set of binary decision variables (spins) $\sets = \{s_i|s_i \in
\{0,1\},i=1,\dots,N\}$. If we assume a Boltzmann probability
distribution for the spins, the average $\avsi_T$ will be given by
\beq
        \avsi_T = \frac{\displaystyle{\sum_{\sets}} s_i \exp\( - E(\sets)/T\) }
        {\displaystyle{\sum_{\sets}}\exp\( - E(\sets)/T\) }
\eeq
where the sums run over all possible configurations $\sets$. We can
manipulate this expression to obtain,
\beq
        \avsi_T =
        \frac{\displaystyle{\sum_{\sets}} \exp\( - E(\sets)/T\)
        \displaystyle{
        \frac{\displaystyle{\sum_{s_i = 0,1}}
        s_i\exp\( - E(\setsi,s_i)/T\) }
        {\displaystyle{\sum_{s_i = 0,1}} \exp\( - E(\setsi,s_i)/T\) }}}
        {\displaystyle{\sum_{\sets}} \exp\( - E(S)/T\) }
\eeq
where $\setsi$ denotes the set $\{s_j,j\neq i\}$ of all spins but $s_i$. If
we now perform the sums over $s_i$ in the numerator, we get
\beqa
\label{A3}
        \avsi_T &=&
        \frac{\displaystyle{\sum_{\sets}} \exp\( - E(\sets)/T\)
        \displaystyle{
        \frac{1}{1+\exp\( \Delta E_i/T\) }}}
        {\displaystyle{\sum_{\sets}} \exp\( - E(\sets)/T\) }
\quad\equiv\quad
        \left\langle \frac{1}{1+\exp\( \Delta E_i/T\) } \right\rangle_T\;
\\
        \textrm{where} \qquad \Delta E_i(\setsi) &=& E(\setsi,s_i=1)-E(\setsi,s_i=0)
\eeqa

So far there are no approximations, the expression for $\avsi_T$ has
just been rewritten. Eq. (\ref{A3}) states that the expectation value
of $s_i$ is equal to the expectation value of a nonlinear function $f$
of all the other spins. The mean field approximation consists of
approximating the expectation value $\langle f\(\setsi\)\rangle$ by
$f\(\left\langle \setsi \right\rangle \)$. With $v_i$ denoting
$\avsi$, and $\setvi$ the complementary set $\{v_j, j\neq i\}$, this
amounts to making the replacement
\beq
        \left\langle \frac{1}{1 + \exp\(\Delta E_i\(\setsi\)/T\)} \right\rangle_T
        \to
        \frac{1}{ 1 + \exp\(\Delta E_i\(\setvi\)/T\)}
\eeq
in Eq. (\ref{A3}).
This results in a set of self-consistency equations for $\setv$, the
{\em mean field equations}
\beq
        v_i = \frac{1}{1 + \exp\( \Delta E_i\(\setvi\) / T \) },\;\; i=1\dots N
\eeq
which in general must be solved numerically, e.g. by iteration.

The mean field approximation often becomes exact in the limit of
infinite range interactions where each spin variable interacts with
all the others. This can be seen from $\Delta E_i(\setsi)$ which then
becomes a sum of many (approximately) independent random numbers, and
a central limit theory can be applied. 
\newpage

\app{Algorithm Details}
Here follows a summary of the MF annealing algorithm for finding
approximate solutions to set covering problems. The procedure
presented below is used for all problems in this study except for the
large rail problems where numerical problems caused by limited machine
precision comes into play. The problem arises when calculating the
product in Eq. (\ref{dE}), which for large problems can contain many
factors. A work-around is implemented by a simple truncation,
\beq
        v_i := \left\{ \begin{array}{ll}
        0 & \quad\mbox{if $v_i < 0.05$} \\
        v_i & \quad\mbox{otherwise}
        \end{array}
        \right.
\eeq
This numerical fix is only used when calculating the $\Delta
E_i\(\setvi\)$ in Eq. (\ref{dE}).

Algorithmic outline of our approach:
\begin{figure}[htb]
\[\fbox{
\begin{minipage}[t]{10.0cm}
\begin{list}{}{\setlength{\itemsep}{2 pt}}
\item[1.] {Rescale all weights such that $c_j \in [0,1] \quad j=1,\dots,N$}
\item[2.] {Initiate all $v_j$'s close to $0.5$ \\
        ($v_j \in [0.499,0.501]$  uniformly random)}
\item[3.] {Set $\alpha = 1.05 \, c_{max}$, or $\alpha = 0.5$ for unicost problems}
\item[4.] {Set the temperature $T = 2 \, T_c$, or $T = 50$ for unicost problems}
\item[5.] {Randomly (without replacement) select one variable $v_k$}
\item[6.] {Update $v_k$ according to Eq. (\ref{update})}
\item[7.] {Repeat points 5.--6. $N$ times (so that all $v_j$ have been
        updated once)}
\item[8.] {Repeat points 5.--7. until no changes occur \\
        (e.g. defined by $1/N\sum_j^N |v_j-v_j^{(old)}| \leq 0.01$)}
\item[9.] {Decrease the temperature, $T \rightarrow 0.80 T$}
\item[10.] {Repeat 5-9 until $\Sigma$ (Eq. (\ref{sat})) is 
        close to 1 \\
        (e.g. $\Sigma \geq (N-0.5)/N$)}
\item[11.] {Finally, the mean field solution is given by setting \\
        $s_j = 1$ if $v_j \geq 0.5$ and $s_j = 0$ otherwise, $j=1,\dots,N$}
\end{list}
\end{minipage}
}\]
\caption{Implementation details of the mean field algorithm for set
        covering problems.}
\label{MFlg}
\end{figure}

\newpage

\app{Benchmark Results}

\begin{table}[hb]
\begin{center}
\begin{tabular}{lrrccc}
\hline
Problem & Rows & columns & Density & Number of ones per & Number of \\
set & $(M)$ & $(N)$ & (\%) & row [Max-Min-Ave] & Problems \\
\hline
4        & 200  & 1000    & 2    & 36-8-20      & 10 \\
5        & "    & 2000    & 2    & 60-21-40     & " \\
6        & "    & 1000    & 5    & 71-29-50     & 5 \\
A        & 300  & 3000    & 2    & 81-38-60     & " \\
B        & "    & "       & 5    & 191-114-150  & " \\
C        & 400  & 4000    & 2    & 105-56-80    & " \\
D        & "    & "       & 5    & 244-159-200  & " \\
NRE      & 500  & 5000    & 10   & 561-444-499  & " \\
NRF      & "    & "       & 20   & 1086-914-999 & " \\
NRG      & 1000 & 10000   & 2    & 258-153-199  & " \\
NRH      & "    & "       & 5    & 580-436-499  & " \\
Rail507  & 507  & 63009   & 1.3  & 7753-1-807   & 1 \\
Rail516  & 516  & 47311   & 1.3  & 7805-1-610   & " \\
Rail582  & 582  & 55515   & 1.2  & 8919-1-690   & " \\
Rail2536 & 2536 & 1081841 & 0.40 & 86666-1-4335 & " \\
Rail2586 & 2586 & 920683  & 0.34 & 72553-1-3097 & " \\
Rail4284 & 4284 & 1092610 & 0.24 & 56181-1-2633 & " \\
Rail4872 & 4872 & 968672  & 0.20 & 69708-1-1897 & " \\
\hline 
E & 50 & 500 & 20 & 124-77-100 & 5 \\
CYC6 & 240 & 192 & 2.1 & 4-4-4 & 1 \\
CYC7 & 672 & 448 & 0.9 & '' & '' \\
CYC8 & 1792 & 1024 & 0.4 & '' & '' \\
CYC9 & 4608 & 2304 & 0.2 & '' & '' \\
CYC10 & 11520 & 5120 & 0.08 & '' & '' \\
CYC11 & 28160 & 11264 & 0.04 & '' & '' \\
CRL10 & 511 & 210 & 12 & 126-10-26 & '' \\
CRL11 & 1023 & 330 & '' & 210-20-41 & '' \\
CRL12 & 2047 & 495 & '' & 330-30-62 & '' \\
CRL13 & 4095 & 715 & '' & 495-50-89 & '' \\
STS45 & 330 & 45 & 6.7 & 3-3-3 & '' \\
STS81 & 1080 & 81 & 3.7 & '' & '' \\
STS135 & 3015 & 135 & 2.2 & '' & '' \\
STS243 & 9801 & 243 & 1.2 & '' & '' \\
\hline
\end{tabular}
\end{center}
\caption{Test problem details. The density refers to the percentage
of ones in the $A$ matrix. The maximum, minimum and avearage of ones 
per row is taken over all problems in each set. E -- STS 
are unicost problems.}
\label{pdef}
\end{table}

\begin{table}
\small
\begin{center}
\begin{tabular}{lrrrc}
\hline
Problem & Optimal value & Best MF  & Average MF & Solution time \\
& & solution in & solution over & \\
& & 10 trials & 10 trials & \\
\hline
4.1  & 429 & 435 & 435.6 & 0.44 \\
4.2  & 512 & 517 & 518.0 & 0.49 \\
4.3  & 516 & 531 & 532.7 & 0.45 \\
4.4  & 494 & 512 & 520.9 & 0.48 \\
4.5  & 512 & 522 & 524.1 & 0.46 \\
4.6  & 560 & 566 & 567.8 & 0.44 \\
4.7  & 430 & 446 & 446.0 & 0.45 \\
4.8  & 492 & 492 & 493.8 & 0.46 \\
4.9  & 641 & 658 & 661.4 & 0.47 \\
4.10 & 514 & 521 & 521.0 & 0.45 \\

5.1  & 253 & 260 & 268.6 & 0.87 \\
5.2  & 302 & 316 & 316.0 & 0.83 \\
5.3  & 226 & 229 & 229.0 & 0.86 \\
5.4  & 242 & 247 & 247.5 & 0.86 \\
5.5  & 211 & 214 & 214.3 & 0.83 \\
5.6  & 213 & 213 & 213.2 & 0.82 \\
5.7  & 293 & 304 & 305.0 & 0.86 \\
5.8  & 288 & 299 & 300.1 & 0.90 \\
5.9  & 279 & 281 & 281.0 & 0.82 \\
5.10 & 265 & 273 & 274.0 & 0.83 \\

6.1  & 138 & 143 & 143.0 & 0.63 \\
6.2  & 146 & 153 & 153.2 & 0.62 \\
6.3  & 145 & 150 & 150.2 & 0.60 \\
6.4  & 131 & 132 & 133.1 & 0.62 \\
6.5  & 161 & 169 & 169.8 & 0.62 \\

A.1  & 253 & 260 & 261.5 & 1.5  \\
A.2  & 252 & 257 & 258.3 & 1.5  \\
A.3  & 232 & 238 & 241.3 & 1.5  \\
A.4  & 234 & 238 & 239.7 & 1.5  \\
A.5  & 236 & 238 & 238.9 & 1.4  \\

B.1  & 69  & 70  & 71.2  & 2.2  \\
B.2  & 76  & 77  & 77.6  & 2.3  \\
B.3  & 80  & 83  & 83.7  & 2.2  \\
B.4  & 79  & 80  & 80.0  & 2.3  \\
B.5  & 72  & 72  & 72.0  & 2.3  \\

C.1  & 227 & 233 & 233.6 & 2.2  \\
C.2  & 219 & 222 & 224.3 & 2.2  \\
C.3  & 243 & 249 & 251.1 & 2.3  \\
C.4  & 219 & 220 & 220.1 & 2.3  \\
C.5  & 215 & 219 & 219.1 & 2.2  \\

D.1  & 60  & 64  & 64.6  & 3.7  \\
D.2  & 66  & 66  & 66.3  & 3.8  \\
D.3  & 72  & 73  & 75.1  & 3.9  \\
D.4  & 62  & 63  & 63.0  & 3.8  \\
D.5  & 61  & 64  & 64.6  & 3.8  \\
\hline
\end{tabular}
\end{center}
\caption{Results for problems 4 -- D (see Table \protect{\ref{pdef}}).
Solution time refers 400 MHz Pentium II CPU seconds and includes the prerun
for non-unicost problems.}

\label{res1}
\end{table}
\normalsize

\begin{table}
\small
\begin{center}
\begin{tabular}{lrrrc}
\hline
Problem & Current & Best MF  & Average MF & Solution time \\
& best value & solution in & solution over & \\
& & 10 trials & 10 trials & \\
\hline
NRE.1 & 29  & 29  & 29.5  & 9.8 \\
NRE.2 & 30  & 32  & 32.1  & 9.8 \\
NRE.3 & 27  & 28  & 28.2  & 9.7 \\
NRE.4 & 28  & 29  & 29.7  & 9.8 \\
NRE.5 & 28  & 29  & 29.0  & 9.8 \\
NRF.1 & 14  & 14  & 14.9  & 19  \\
NRF.2 & 15  & 15  & 15.4  & 18  \\
NRF.3 & 14  & 15  & 15.2  & 19  \\
NRF.4 & 14  & 15  & 15.4  & 19  \\
NRF.5 & 13  & 14  & 14.7  & 19  \\
NRG.1 & 176 & 180 & 180.1 & 10  \\
NRG.2 & 155 & 157 & 159.0 & 10  \\
NRG.3 & 166 & 173 & 174.9 & 10  \\
NRG.4 & 168 & 175 & 176.3 & 10  \\
NRG.5 & 168 & 175 & 176.9 & 11  \\
NRH.1 & 64  & 65  & 66.4  & 21  \\
NRH.2 & 64  & 66  & 67.0  & 21  \\
NRH.3 & 59  & 62  & 62.8  & 20  \\
NRH.4 & 58  & 60  & 61.8  & 21  \\
NRH.5 & 55  & 56  & 56.4  & 21  \\
Rail507  & 174  & 187  & 188.2   & 37  \\
Rail516  & 211  & 186  & 187.9   & 26  \\
Rail582  & 182  & 222  & 225.5   & 32  \\
Rail2536 & 691  & 737  & 740.0   & 1100 \\
Rail2586 & 951  & 1018 & 1026.7  & 830  \\
Rail4284 & 1065 & 1152 & 1162.1  & 1100 \\
Rail4872 & 1534 & 1640 & 1643.5  & 1050 \\
\hline
\end{tabular}
\end{center}
\caption{Results for problems NRE -- Rail (see Table \protect{\ref{pdef}}).
Solution time refers 400 MHz Pentium II CPU seconds and includes the prerun for
non-unicost problems. The notation current best 
value may for some instances mean optimal value.}
\label{res2}
\end{table}
\normalsize

\begin{table}
\small
\begin{center}
\begin{tabular}{lrrrc}
\hline
Problem & Current & Best MF  & Average MF & Solution time \\
& best value & solution in & solution over & \\
& & 10 trials & 10 trials & \\
\hline
E.1     & 5    & 5    & 5.3    & 0.15   \\
E.2     & 5    & 5    & 5.0    & 0.14   \\
E.3     & 5    & 5    & 5.0    & 0.15   \\
E.4     & 5    & 5    & 5.0    & 0.14   \\
E.5     & 5    & 5    & 5.0    & 0.16   \\
CYC.6   & 60   & 62   & 63.0   & 0.08   \\
CYC.7   & 144  & 151  & 153.4  & 0.20   \\
CYC.8   & 344  & 348  & 352.1  & 0.62   \\
CYC.9   & 780  & 829  & 832.6  & 1.6    \\
CYC.10  & 1792 & 1870 & 1882.3 & 3.9    \\
CYC.11  & 4103 & 4240 & 4248.7 & 9.6    \\
CLR.10  & 25   & 27   & 29.0   & 0.36   \\
CLR.11  & 23   & 26   & 28.9   & 1.0    \\
CLR.12  & 26   & 30   & 30.9   & 3.2    \\
CLR.13  & 26   & 31   & 32.9   & 10     \\
STS.45  & 30   & 31   & 31.8   & 0.03   \\
STS.81  & 61   & 63   & 63.9   & 0.11   \\
STS.135 & 104  & 105  & 107.4  & 0.32   \\
STS.243 & 202  & 205  & 206.8  & 1.1    \\
\hline
\end{tabular}
\end{center}
\caption{Results for the unicost problems E -- STS 
(see Table \protect{\ref{pdef}}). 
Solution time refers 400 MHz Pentium II CPU seconds. The notation current best 
value may for some instances mean optimal value.}
\label{res3}
\end{table}
\normalsize

\newpage

\FloatBarrier
\app{WWW Server}

A program executing the mean field algorithm for set covering problems
as presented in this paper can be publicly used by means of a World
Wide Web server. A user can interactively submit a file defining an
instance of SCP, and obtain the found solution. The URL of the WWW
server is:
\begin{center}
{\tt http://www.thep.lu.se/complex/mf\_server.html}
\end{center}
An instance of SCP is defined by specifying the costs $c_i$ and the
matrix $A$. Two formats, row and column ordering, are supported for
the file that lists $c_i$ and the none-zero entries of $A$; they are
defined as follows:
\begin{table}[h]
\begin{center}
\begin{tabular}{l}
{\bf Row Ordering}\\
\hline
$M$ $N$  \\
$c_1 c_2 \dots c_N$  \\
$\hat{M}_1$   ``space separated list of all non-zero entries for row 1'' \\
$\hat{M}_2$   ``space separated list of all non-zero entries for row 2'' \\
\dots  \\
$\hat{M}_M$   ``space separated list of all non-zero entries for row M'' \\
\hline
\\
{\bf Column Ordering}\\
\hline
$M$ $N$ \\
$c_1$ $\hat{N}_1$  ``space separated list of all non-zero entries for column 1'' \\
$c_2$ $\hat{N}_2$  ``space separated list of all non-zero entries for column 2'' \\
\dots \\ 
$c_N$ $\hat{N}_N$  ``space separated list of all non-zero entries for column N'' \\
\hline
\end{tabular}
\end{center}
\end{table}

The column sums $\hat{N}_i$ and row sums $\hat{M}_k$ are defined in
Eqs. (\ref{colsum}-\ref{rowsum}).
As an example, consider the SCP instance defined by  
\beq \vec{c} = (1,2,3,4,5),\quad A = \left(
   \begin{array}{ccccc}
        1 & 0 & 1 & 0 & 1 \\
        0 & 1 & 0 & 1 & 0 \\
        1 & 1 & 0 & 0 & 1 \\
        0 & 0 & 1 & 1 & 1 \\
   \end{array}
\right)
\eeq
for which the column and row ordering formats read:
\begin{center}
\parbox[t]{5cm}{
\textbf{Row Ordering}
\vspace{0.1in}\\
4 5 \\
1 2 3 4 5 \\
3 1 3 5 \\
2 2 4 \\
3 1 2 5 \\
3 3 4 5 \\
}
\parbox[t]{5cm}{
\textbf{Column Ordering}
\vspace{0.1in}\\
4 5 \\
1 2 1 3 \\
2 2 2 3 \\
3 2 1 4 \\
4 2 2 4 \\
5 3 1 3 4 \\
}
\end{center}

The solver returns the found cost (energy) $E$ (Eq.(\ref{def})),
together with a characterization of the problem. Upon request, a file
that lists the columns used in the solution is also provided.

There is a limitation ($B$) on the size of the problems that can be
submitted to the server. Instances of SCP with $MN\rho > B$ will not
be considered. Presently $B$, which is limited by the available memory
of the server, is given by $3x10^6$.


\begin{thebibliography}{99}

\bibitem{beas1} J. E. Beasley,
``An Algorithm for the Set Covering Problem'',
{\it European Journal for Operations Research} 31 (1987) 85-93.

\bibitem{beas2} J. E. Beasley,
``A Genetic Algorithm for the Set Covering Problem'',
{\it European Journal for Operations Research} 94 (1994) 392-404.

\bibitem{sass1} S. Ceria, P. Nobili and A. Sassano,
``A Lagrangian-Based Heuristics for Large Scale Set-Covering Problems'',
\textit{Mathematical Programming B} 81 (1998).

\bibitem{gross} T. Grossmann and A. Wool, 
``Computational Experience with Approximation Algorithms for 
the Set Covering Algorithm'', 
{\it European Journal for Operations Research} 101 (1987) 81-92.

\bibitem{tank} J.J. Hopfield and D.W. Tank,
``Neural Computation of Decisions in Optimization Problems'',
{\it Biological Cybernetics} 52 (1985) 141-152.

\bibitem{lager1} J. H\"akkinen, M. Lagerholm, C. Peterson and
B. S\"oderberg, ``A Potts Neuron Approach to Communication Routing''
{\it Neural Computation} 10 (1998) 1587-1599.

\bibitem{res} N. Karmarkar, M.G.C. Resende and K.G. Ramakrishnan,
``An Interior Point Algorithm to Solve Computationally 
Difficult Set Covering Problems'', \textit{Mathematical Programming} 
52 (1991) 597-618.

\bibitem{kirk} S. Kirkpatrick, C.D. Gelatt and M.P. Vecchi,
``Optimization by Simulated Annealing'',
{\it Science} 220 (1983) 671.

\bibitem{lager} M. Lagerholm,  C. Peterson and B.  S\"oderberg,
``Airline Crew Scheduling with Potts Neurons'',
{\it Neural Computation} 9 (1997) 1589-1599.

\bibitem{lager2} M. Lagerholm,  C. Peterson and B.  S\"oderberg,
``Airline Crew Scheduling Using Potts Mean Field Techniques'',
{\it LU TP 97-10}
(to appear in {\it European Journal for Operations Research}).

\bibitem{sass2} C. Mannino and A. Sassano,
``Solving Hard Set Covering problems''
\textit{Operations Research Letters} 18 (1995) 1-5. 

\bibitem{mo1} M. Ohlsson and H. Pi,
``A Study of the Mean Field Approach to Knapsack Problems'',
{\it Neural Networks} 10 (1997) 263-271.

\bibitem{ohl} M. Ohlsson, C. Peterson and B. S\"oderberg,
``Neural Networks for Optimization Problems with Inequality
Constraints - the Knapsack Problem'',
{\it Neural Computation} 5 (1993) 331-339.

\bibitem{OR} The OR-Library, {\tt http://www.ms.ic.ac.uk/info.html}.

\bibitem{parisi} G. Parisi, {\it Statistical Field Theory},
Addison-Wesley, 1988.

\bibitem{pet1} C. Peterson and B. S\"oderberg,
``A New Method for Mapping Optimization Problems onto
Neural Networks'',
{\it International Journal of Neural Systems} 1 (1989) 3-22.

\end{thebibliography}
\end{document}